%% file: 0-main.tex
\documentclass[letterpaper, 10 pt, conference]{ieeeconf}
\IEEEoverridecommandlockouts
\usepackage{times}
\usepackage{bm}
\usepackage{soul}
\usepackage{amsfonts}
\usepackage{multicol}
\usepackage[hidelinks]{hyperref}
\usepackage{booktabs} 
\usepackage[table,dvipsnames]{xcolor}
\let\labelindent\relax
\usepackage{enumitem}
\usepackage{graphicx}
\usepackage{grffile}
\usepackage[font={small,it}]{caption}
\usepackage{algorithm}
\usepackage{algpseudocode}
\usepackage{caption} 
\usepackage{subcaption}
\usepackage[para,online,flushleft]{threeparttable}
\usepackage{balance}
\usepackage{svg}
\usepackage{float}
\usepackage{siunitx}

\begin{document}

\title{Elly: A Real-Time Failure Recovery and Data Collection\\System for Robotic Manipulation}


\author{Elena Galbally\textbf{*}, Adrian Piedra\textbf{*}, Cynthia Brosque, and Oussama Khatib\\
\thanks{\textbf{*} Both authors contributed equally to the paper. All authors are part of the Stanford Robotics Lab. Contact: {\tt\small elenagal@stanford.edu} or {\tt\small apiedra@stanford.edu}.}}




\maketitle

\begin{abstract} 
Even the most robust autonomous behaviors can fail. The goal of this research is to both recover and collect data from failures, during autonomous task execution, so they can be prevented in the future. We propose haptic intervention for real-time failure recovery and data collection. Elly is a system that allows for seamless transitions between autonomous robot behaviors and human intervention while collecting sensory information from the human’s recovery strategy. The system and our design choices were experimentally validated on a single arm task -- installing a lightbulb in a socket -- and a bimanual task -- screwing a cap on a bottle -- using two 7-DOF manipulators equipped 4-finger grippers. In these examples, Elly achieved over 80\% task completion during a total of 40 runs. 
\end{abstract}

\input{1-Introduction.tex}
\input{2-RelatedWork.tex}

\input{3-Methods.tex}

\input{4-Experiments.tex}
\input{5-Conclusions.tex}


\bibliographystyle{plain}
\footnotesize
\bibliography{references}

\end{document}

%% file: 1-Introduction.tex
\section{Introduction}
\label{sec:intro}

Humans are extremely creative. Over time, we have designed myriad tools and objects to help with our daily tasks. This diversity makes human environments exciting yet challenging for automation. The amount of uncertainty autonomous systems need to face makes it difficult to guarantee task success and safety. In fact, no fully autonomous system is completely bulletproof. Even vision systems with access to huge data sets like ImageNet \cite{Deng2010ImageNet:Database} will sometimes mistake a cat for a cow. Embodied autonomous systems, such as robots, have a more challenging task: completing physical actions that interact with an uncertain world, all based on scarce training data.


Since failures -- that is, states that prevent task completion because the robot does not know how to proceed - are bound to occur, we need to design systems that can recover and learn from mistakes. Furthermore, the behaviors learned should be modular so that the robot can generalize to new tasks without having to learn the entire behavior from scratch. Throughout the paper, we will refer to these modular behaviors as primitives. In this work, we design and test Elly -- a system that uses human data and physical priors to learn autonomous manipulation tasks. Furthermore, Elly allows a human operator to intervene in the event of failure and collects data from the failure recovery strategy to improve the robot's knowledge of the task.


\begin{figure}[ht!]
    \centering
    \includegraphics[width = \columnwidth]{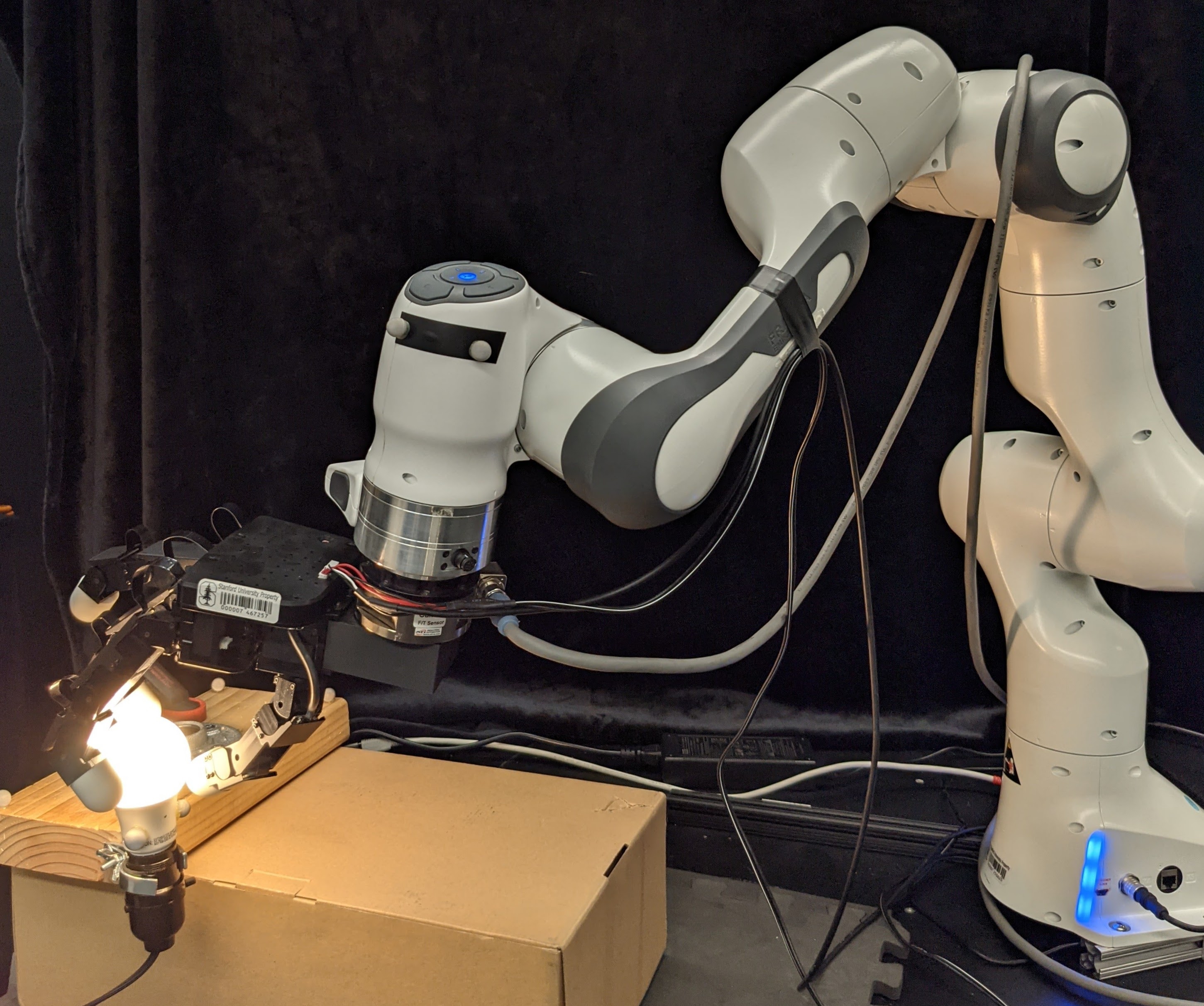}
    \caption{Lightbulb installation task performed using a 7-DOF arm equipped with a 4 finger gripper. The system presented in this paper was also tested on a bimanual task. For more experimental results and implementation details checkout the \href{https://egalbally.github.io/learning\_skills/}{\underline{project website}}.}
    \label{fig:abstract}
\end{figure}


Our system design decisions were guided by six high-level characteristics described here:


\textbf{Safe.} When operating with and around humans, safety is of the utmost importance. Hence, we cannot define our autonomous primitives as pre-programmed trajectories and execute them using open-loop control. Additionally, in order to prevent unsafe behaviors, we want the primitive outcomes to be predictable and their representation to be physically meaningful. Model-based controllers parametrized using human demonstration data enable us to work in a low-dimensional latent space and are generally easier to interpret than data-driven approaches. 

\textbf{Capable of Recovering from Failures.} When reaching a failure state, a human operator should be able to intervene in order to complete the task. In turn, this intervention data should be collected to improve the robot's knowledge of the task. Over time, the idea is to have the robot become less dependent on human intervention. For applications such as construction tasks, keeping the human in the loop has the added benefit of preserving jobs. Given we are dealing with contact-rich tasks, a haptic device is the most appropriate interface since it provides force feedback to the operator.

\textbf{Hardware-driven.} Solutions should work on real hardware. This means that we prioritized using methods and algorithms that run reliably on our 7 degree of freedom (DOF) robots. There exist more cutting-edge algorithms for perception, haptic and autonomous control, however, there is value in identifying a set of ``simple'' algorithms that are sufficiently robust to allow testing of the complete system. This allowed us to identify cross-cutting issues that only reveal themselves when building an end-to-end system. We hope these insights will be valuable for future research.

\textbf{Data-efficient.} Simulation environments provide a valuable way to collect large amounts of data quickly. However, accurate contact rendering is a well-known limitation of most physics simulators and an ongoing area of research \cite{Erez2015SimulationPhysX, Liu2008ASurfaces, Zhu2019AnAssemblages}. Furthermore, behaviors learned in simulation do not always transfer to the real world. Thus, simulated data is not a good option for our experiments. However, since collecting demonstrations on real hardware is time-consuming, the algorithm choice is constrained by the amount of data available. Thus, we need to use data-efficient algorithms and low-dimensional representations. 



\textbf{Modular Behaviors.} We do not want to program new tasks entirely from scratch. Instead, we want to break the tasks down into primitives that can be reused to perform new tasks. In this paper, we reuse the same 5 primitives -- go-to-point, make contact, align, engage on threads, and screw -- to perform two different tasks: screwing a cap on a bottle and installing a light bulb in its socket. 

\textbf{Object-centric.} Elly should generalize to different robotic platforms. To do that, we need to encode the primitive behaviors to focus on what is happening to the object instead of what the robot is doing \cite{Migimatsu2020Object-CentricEnvironments}. The behavior of an object can be fully characterized by a frame \cite{Sharma2013GeneralizingKeypoints}, that is, an origin and 3 directions in space about which we control forces and moments, $\bm{F}$. The Operational Space formulation \cite{Khatib1987AFormulation} is an object-level control framework for torque-controlled robots that uses precise models of the robot to map our desired behavior at the object frame $\bm{F_o}$ to the required robot torques at the joints $\bm{\tau} = J^T(\bm{q})\bm{F_o}$.
    

\vspace{7pt}
This paper focuses on the system components that most strongly relate to our primary goals while making assumptions about others. For instance, we use motion capture (mocap) to get rough estimates of our desired target locations during free space motion rather than diving into the problem of object recognition. We also assume that the object is grasped in such a way that the assembly task is possible \cite{Speziali2020RobustHand}. For example, we do not consider cases in which the light bulb would be grasped by its threads. 



In summary, our key contributions are:
\begin{itemize}
    \item Flexible control framework that combines robot-agnostic autonomous primitives with real-time haptic intervention for failure recovery and data collection.
    \item Complete documentation and resources to replicate our hardware setup for bimanual human-robot collaboration and data collection using haptic feedback\footnote{https://egalbally.github.io/LearningRobotSkills/}.
    \item Experimental results that validate our system and design choices on two contact tasks: screwing a lightbulb into a fixed socket using one 7-DOF arm and screwing a cap on a bottle using two 7-DOF manipulators.
\end{itemize}

In addition, our code and datasets are publicly available\footnote{https://github.com/egalbally/LearningRobotSkills}. They comprise the force sensor, haptic, and mocap sensory information collected during our experiments. More broadly, we hope other researchers will benefit from this work by: using our dataset to train their own models to learn contact behaviors in novel ways; replicating our hardware setup and adapting our graphical user interface (GUI) to collect their own datasets of interest; or incorporating haptics into their control framework to intervene during failures and ensure task completion.

\begin{figure*} [ht!]
    \centering
    \includegraphics[width = \textwidth]{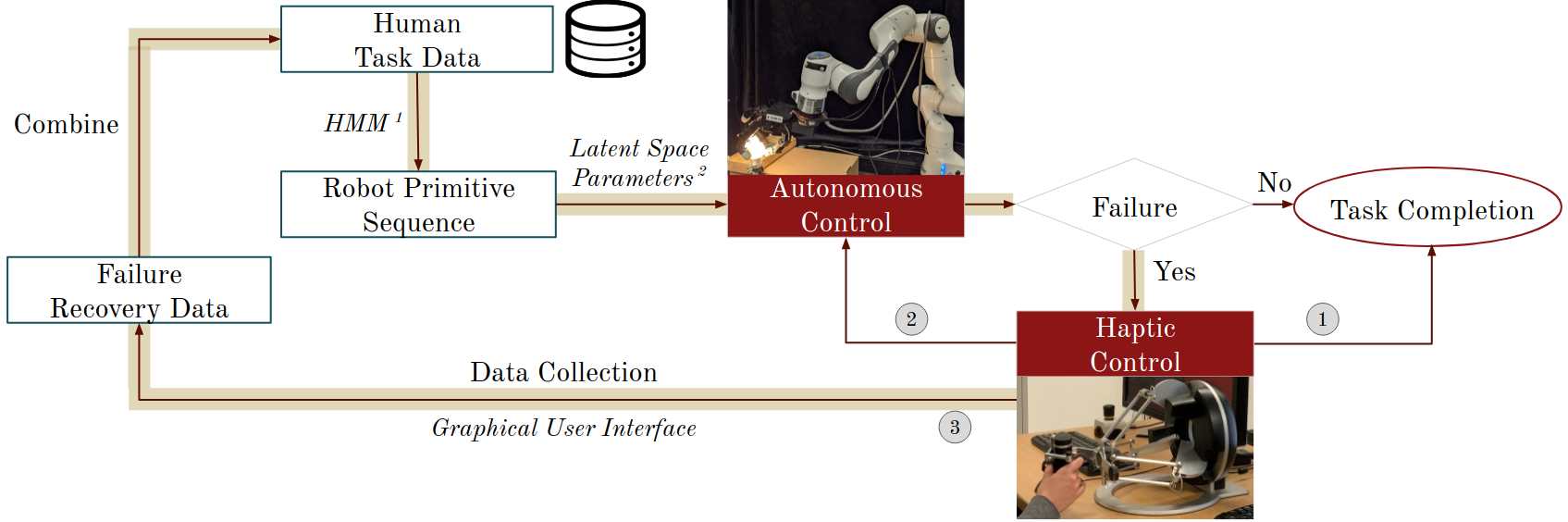}
    \caption{System Diagram: We collect a small set of human demonstrations of a complete task. The demonstrations are then segmented into a sequence of primitives and used to parametrize the autonomous model-based controllers. In case of failure, we switch to haptic control which enables task completion while providing the operator with real-time contact feedback. Data from the failure recovery strategies is collected and added to the original data set to improve the robot's knowledge of the task. The yellow highlight outlines the interactive haptic learning loop. We built upon prior work ( $^1$\cite{Herrero2021UnderstandingPrimitives} and $^2$\cite{GalballyHerrero2022ParametrizationDemonstrations}) to parametrize the autonomous controllers and build this end-to-end system that was experimentally evaluated on real hardware.}
     \label{fig:concept_diagram}
\end{figure*}

%% file: 2-RelatedWork.tex
\section{Related Work}
\label{sec:related}

The different elements of our system can fall under three broad categories: autonomous manipulation primitives, learning, and human-robot interfaces. Here we will provide a brief overview of related work and alternative approaches in each of these areas of research.

\textbf{Primitives.} We use primitives as our building blocks for the autonomous task execution. They allow us to plan at a higher level of abstraction \cite{Migimatsu2021GroundingActions} than. In particular, we parametrize our primitive controllers using the notion of frames. Prior literature \cite{Sharma2013GeneralizingKeypoints} uses this same notion of object-centric, task-axes controllers. In contrast to the work presented in this paper,  \cite{Sharma2013GeneralizingKeypoints} extracts the controller parameters exclusively from visual data. We believe taking into account force data is advantageous when dealing with contact tasks. Compliance also plays a major role in task success in the presence of position uncertainty or for contact-rich tasks such as grasping or the ones studied in this paper.

The notion of compliant motion primitives has been explored, for instance, in \cite{Kazemi2012RobustPrimitives} where compliant primitives are used to achieve robust grasping of small objects. Examples of learning compliant motions from demonstration can be found in \cite{SuomalainenImitationMotions, Kormushev2011ImitationInput}. In our case, we do not only learn motions, but also desired forces and moments \cite{GalballyHerrero2022ParametrizationDemonstrations}. These values are used to parametrize model-based primitive controllers \cite{Berscheid2021RobotPrimitives, Calinon2018RobotModels}. The low-dimensionality of our models allows us to learn the parameters from a very reduced number of demonstrations. Other work that focuses on data efficient learning includes \cite{Zhan2020AManipulation} and \cite{Deisenroth2015GaussianControl}.

\textbf{Learning.} Humans are extremely capable when it comes to manipulation, so it is no surprise that collecting data from human demonstrations is a popular way to learn new robot behaviors \cite{Langley2021TheMotivation, Yang2019RobotPrimitives, Niekum2013IncrementalDemonstration, Ijspeert2002MovementRobots}. In fact, we can learn not only one arm but also two arm behaviors, as demonstrated by our system and explored in other work \cite{Pairet2019LearningManipulation}. Interactive learning, which keeps human in the loop for continued learning has been a growing area of research in recent years \cite{Hoque2021LazyDAgger:Learning, Macglashan2017InteractiveFeedback, Chisari2022CorrectManipulation, Londono2022DoingCollaboration, Londono2022DoingCollaboration}. In a similar fashion to this body of work, Elly keeps the human in the loop through the use of use of a haptic interface. The failure recovery data collected by Elly could be combined with work such as \cite{Grollman2011DonutDemonstrations} or \cite{Grollman2012RobotDemonstrations} that focuses on how to learn from failures to produce more robust autonomous behaviors. 

\textbf{Human-Robot Interfaces.} 
In order to keep humans in the loop, we need an interface that provides sufficient information for the user to effectively control the robot. The force feedback provided by haptic devices proves extremely helpful for performing contact tasks. An example that uses haptic demonstrations to learn new behaviors can be found in \cite{Calinon2019LearningInterface}. Elly goes beyond only demonstration by enabling haptic intervention during autonomous task execution. This idea of haptic and autonomous control switching has been explored in prior work in the context of shared control \cite{Bustamante2021TowardAssistance} and construction applications \cite{Brosque2021CollaborativeFeedback}. Additionally, advances in the design of affordable and compact haptic devices \cite{Vullez2018LeActifs.} could make haptic research,  more accessible. Our system could easily be adapted to accommodate a different device.

%% file: 3-Methods.tex
\section{Methods} 
\label{sec:methods}

We use haptics for failure recovery, a GUI to collect relevant task data, and build upon prior work to label this data \cite{Herrero2021UnderstandingPrimitives} and parametrize the autonomous model-based primitives \cite{GalballyHerrero2022ParametrizationDemonstrations}. In \cite{Herrero2021UnderstandingPrimitives} we developed a weakly supervised algorithm that automatically labels and segments human demonstrations into sequences of primitives. However, this paper is the first time using this labeler for a real hardware application. 

\subsection{Data Collection - The GUI}
\begin{figure*} [ht]
\centering
    \includegraphics[width = \textwidth]{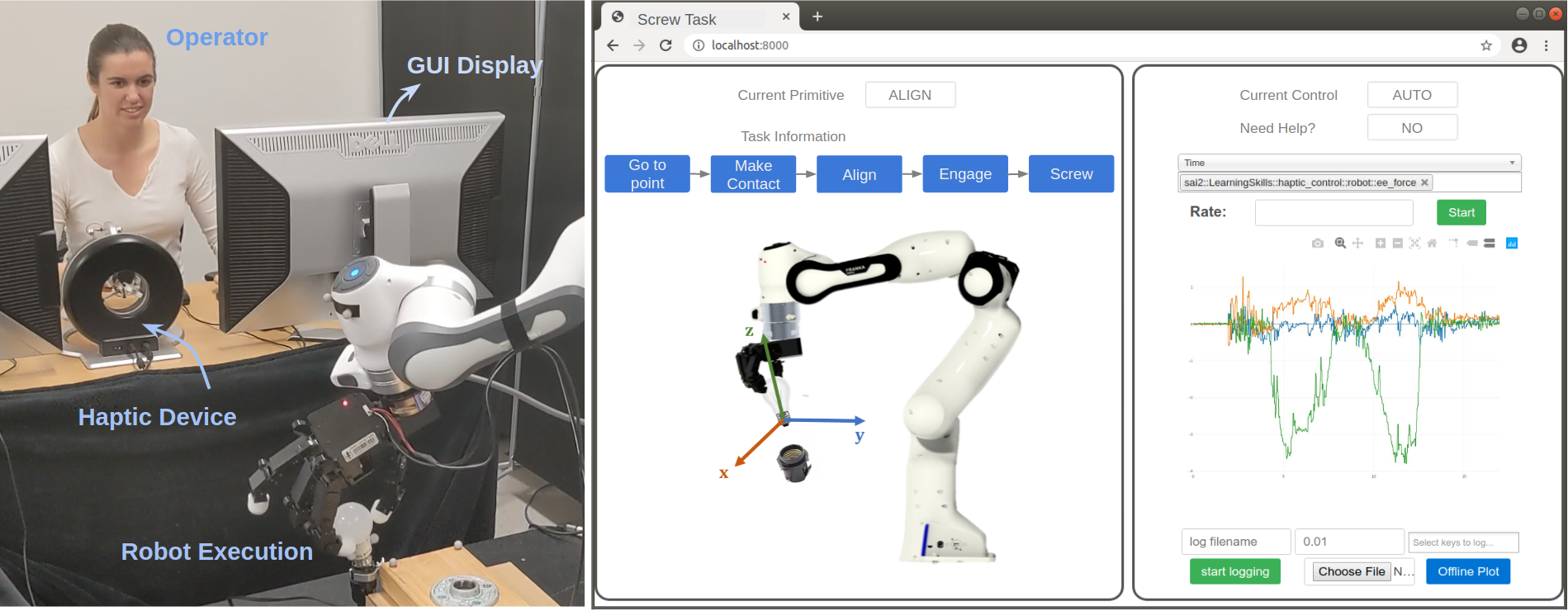}
    \caption{\textbf{Left:} Human operator haptically controlling the robot during a demonstration of the lightbulb task. \textbf{Right:} Graphical User Interface provides the operator with task execution information and a means to collect new data.}
    \label{fig:gui}
\end{figure*}

The first step is data collection. Fig. \ref{fig:gui} illustrates a human-robot collaboration setup and provides a snapshot of the GUI used for data collection. The left panel is an information display, while the right panel is aimed at failure intervention and data logging. The \textbf{information display} shows the current primitive being executed (during ``auto'' mode) and provides information specific to the current task, such as the most common demonstrated sequence of primitives and the location of the control frame being used. The ``current primitive'' variable is automatically updated in real-time by the controller via Redis \cite{RedisBooks}, while we customized the task information to those tackled in this paper. In future work, we would like to automatically update and display the control frame location.

The \textbf{failure intervention and logging} panel automatically updates the control mode (haptic or auto), requests help in the event of failure, and provide a customizable real-time plotter and logger. The ``need help?'' variable will display a \textit{yes} when the autonomous controller has not reached its goal state for that primitive, cannot transition to the next scheduled primitive, and has been in this situation for longer than 15 seconds. It could be possible to learn failure conditions for each primitive based on the data; however, the heuristic used in this paper proved robust enough to ensure task success while avoiding unnecessary intervention requests. Regarding the plotter and logger, all controller and sensor variables are available to the GUI via Redis. Therefore, the operator can choose what variables to save and display. In the case of Fig. \ref{fig:gui} we displayed forces at the end effector and can see that the lightbulb has made contact with the socket twice. 

\subsection{Haptic Control}
 
Haptic devices allow the operator to feel when the robot makes contact with the environment, making them ideal tools for contact-rich assembly tasks such as the ones studied in this paper. A benefit of haptics over direct human demonstrations is that the setup is not task-specific, enabling data collection for a wide variety of single-arm or two-arm manipulation tasks without modifications to the hardware or software setup. Furthermore, haptics operation forces the operator to use and plan with the capabilities and limitations of the robot instead of their own, leading to data close to that encountered during autonomous execution. 

Some of the most common types of haptic controllers include impedance-based \cite{Hogan1984ImpedanceManipulation}, admittance-based \cite{Maples1986ExperimentsManipulators}, hybrid approaches combining the two \cite{Ott2015AControl,Kim2018WeightedInteraction}, and adaptive control laws \cite{Park2016AControl:, Shahdi2009Adaptive/robustTeleoperation}. However, in this work, we choose to use an approach that relies on independent, autonomous behaviors on both the robot and haptic side \cite{Jorda2020RobustInteractions}. This control strategy has shown to be effective at dealing with communication delays and uncertainties.

Another benefit of using haptics is that it can used not only during the initial human demonstration phase of the experiments, but also during failure intervention. We are able to seamlessly switch back and forth between haptic control and autonomous behaviors (see Fig. \ref{fig:switch}) to ensure task completion.

\subsection{Autonomous Control - The Primitives}
We use model-based task space controllers to: ensure robot independent behaviors and reduce the number of parameters we need to learn from our scarce dataset. The Operational Space control framework \cite{Khatib1987AFormulation} is an object-level control framework for torque-controlled robots that relies on precise models of the robot and allows for a dynamically consistent operation of the robot. 

Unified motion and force controllers in operational space require a set of desired forces, moments, positions and orientations which can be extracted from the data. Let the robot have a task to fulfill, described by the task Jacobian $J_t$, the task coordinates $x_t$, and the associated task velocity $\dot{x}_t$, such that $\dot{x}_t = J_t\dot{q}$, where $q$ represents the robot generalized coordinates and $\dot{q}$ represents the robot generalized velocities. The operational space equation of motion is:
\begin{equation}
    \Lambda_t \ddot{x_t}+\mu_t+p_t=F_t \label{eq:motion_op_space}
\end{equation}

\noindent
where $\mu_t = \bar{J_t^T}b - \dot{J_t}\dot{q}$ is the task space Coriolis and centrifugal, and $p_t = \bar{J_t^T}g$ is the gravity projected onto the task space. 
The task control torques will then be $\Gamma_t = J_t^TF_t$. 

\subsection{Regrasping}
Rigidly attaching an object to the end-effector largely simplifies the task by reducing uncertainty from slippage and variations in grasp configuration. Furthermore, developing robust grasping behaviors is not a goal of this work. However, a rigid attachment would not allow us to regrasp objects, which is often required for screwing due to the range limits of the robot joints. For instance, the Panda robots used in our experiments cannot fully screw in a lightbulb and get it to turn on without regrasping along the way. 

Using a gripper allowed us to easily switch between objects during the experiments and makes our system less task-specific. Moreover, it showed that both the haptic and autonomous controllers are robust enough to handle uncertainties in the grasp configuration. The grasp poses were hard coded for each object and regrasping was included as part of the screwing and engaging primitives. Both screwing and engaging involve rotating about an arbitrary axis (CW and CCW, respectively) while applying a downward force along that same axis. Regrasping occurs when the robot is close to a joint limit.

%% file: 4-Experiments.tex
\section{Experiments}
\label{sec:experiments}
This section provides details about the selected tasks and setup, as well as the experimental results and insights. 
\begin{figure*} [ht]
    \centering
    \includegraphics[width = \textwidth]{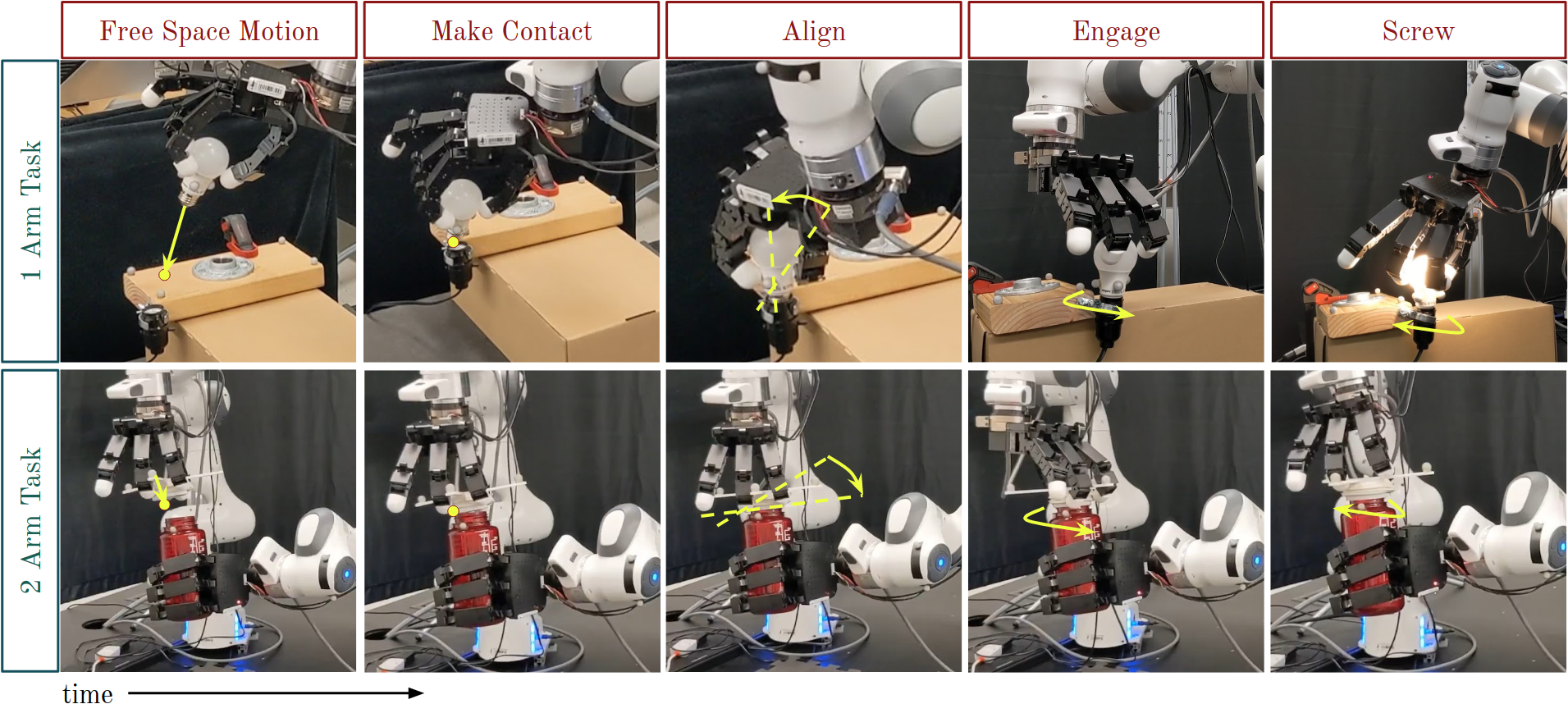}
    \caption{\textbf{Left to right}: Cap on bottle screwing task using a Panda arm, generalization to two arms, and generalization to a different object using the same primitive actions.}
     \label{fig:experiments_robotPics}
\end{figure*}

\subsection{Tasks}
Elly was validated on a lightbulb installation and a bottle cap screwing task. We chose these tasks because they: share primitives, present complex contact states, and provide an opportunity to extend the system to two arms. 

The shared primitives allowed us to test whether our autonomous model-based controller could generalize to new tasks. We require a model of the objects we are manipulating and must calibrate our sensors accordingly. However, we do not need to modify the control laws. For example, pushing down while turning works for screwing both a bulb and a cap. Thus, we trade having to collect new data and treating the two tasks as different problems for requiring a model of the robot and the object. 

Contact is difficult to simulate. Therefore, generating physically-accurate data to learn tasks such as the ones presented here is extremely difficult. In particular, threads present a unique challenge since many simulators do not behave well when dealing with non-convex hull surfaces. Elly allows us to overcome these challenges by enabling real-world data collection. Furthermore, the haptic interface makes it more intuitive for a human to perform these contact tasks. 

Industrial environments often use fixtures to hold objects in place. However, many of our daily tasks do not involve such fixtures \cite{Shao2020LearningSkills}. Placing a lightbulb in a socket is one example of a realistic household task that can be executed using a single arm. Starting with this task gave us the chance to test the basic concepts and functionality of the system before increasing the complexity of the hardware setup. We then moved on to a two-arm bottle task. 

As seen in Table \ref{tab:experiments}, along with the complete execution of the two tasks, we also performed other experiments to illustrate specific aspects or capabilities of the system. In total, we ran 65 experiments on the system, 40 of those using the bimanual setup. 

\begin{table}[!h]
    \caption{Experiments}
    \label{tab:experiments}
    \centering
    \begin{tabular}{lll}
        \toprule
        \textbf{Task} & \textbf{Experiment Focus} & \textbf{\# runs}\\
        \midrule
        Light Bulb & Haptic-Auto Switching  & 5  \\
        Two-arm Bottle & Two-arm Tracking  & 10  \\
        Two-arm Bottle & Regrasping & 10 \\  
        Light Bulb & Task completion  & 20 \\
        Two-arm Bottle & Task completion  & 20 \\
        \bottomrule
    \end{tabular}
\end{table}


\subsection{Hardware Setup}
Elly relies on a complex combination of hardware components that must all communicate with each other. Below is a detailed explanation of the different elements. For further details and resources, such as hardware drivers and CAD models of different adapters, refer to the project website.

\textbf{Manipulator.} We use two 7-DOF torque-controlled Panda arms with a $3\,\mathrm{Kg}$ payload and a maximum reach of $855\,\mathrm{mm}$. Our controllers communicate with the Panda Control Interface over Ethernet at a frequency of $1\,\mathrm{KHz}$. \textbf{Gripper.} The Allegro anthropomorphic hand with four fingers has a $5\,\mathrm{Kg}$ payload and 16-DOF. It communicates over CAN with a $333\,\mathrm{Hz}$ frequency. We chose this hand over a parallel gripper because it can perform pinch and envelope grasps, which are important when manipulating household objects. \textbf{Haptic Device.} We use two Force Dimension haptic devices: the Omega.6 and Omega.7. They both communicate over USB and can render translational forces of up to $12\,\mathrm{N}$. The only difference is that the Omega.7 has an additional analog gripper instead of a button. \textbf{Touch.} We mounted an ATI 6-DOF force/torque sensor to the wrist of each Panda arm. The sensed forces were filtered, scaled, and rendered back to the operator through the haptic device. \textbf{Vision.} We used 10 Prime-13 OptiTrack cameras to provide us with estimates for the object poses. These estimates were used during free space motion to approach a desired object prior to grasping or entering contact.  


\begin{figure*}[t!]
    \centering
    \begin{subfigure}{0.5\textwidth}
      \centering
      \includegraphics[width = 0.95\columnwidth]{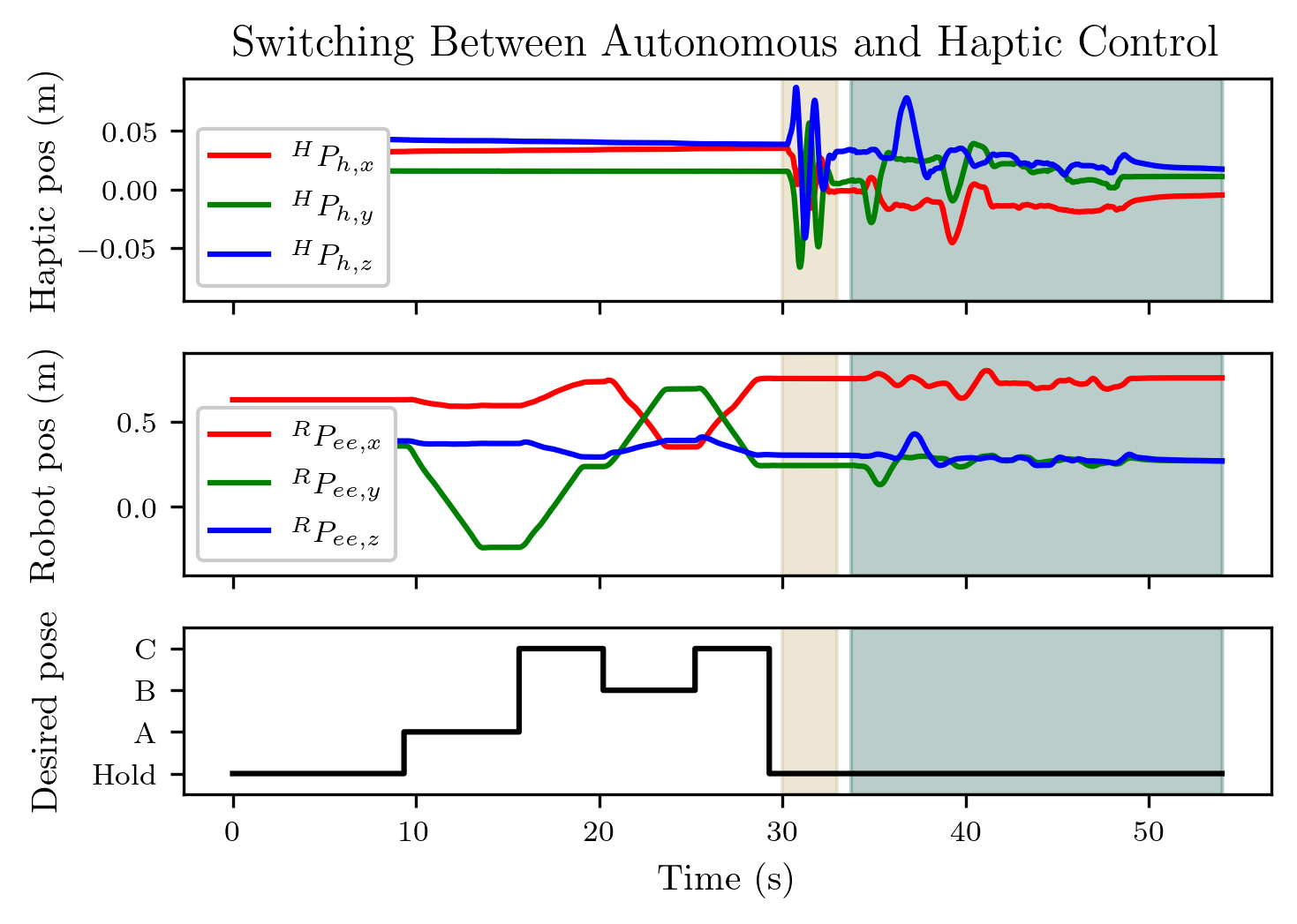}
    \end{subfigure}%
    \begin{subfigure}{0.5\textwidth}
      \centering
      \includegraphics[width = 0.95\columnwidth]{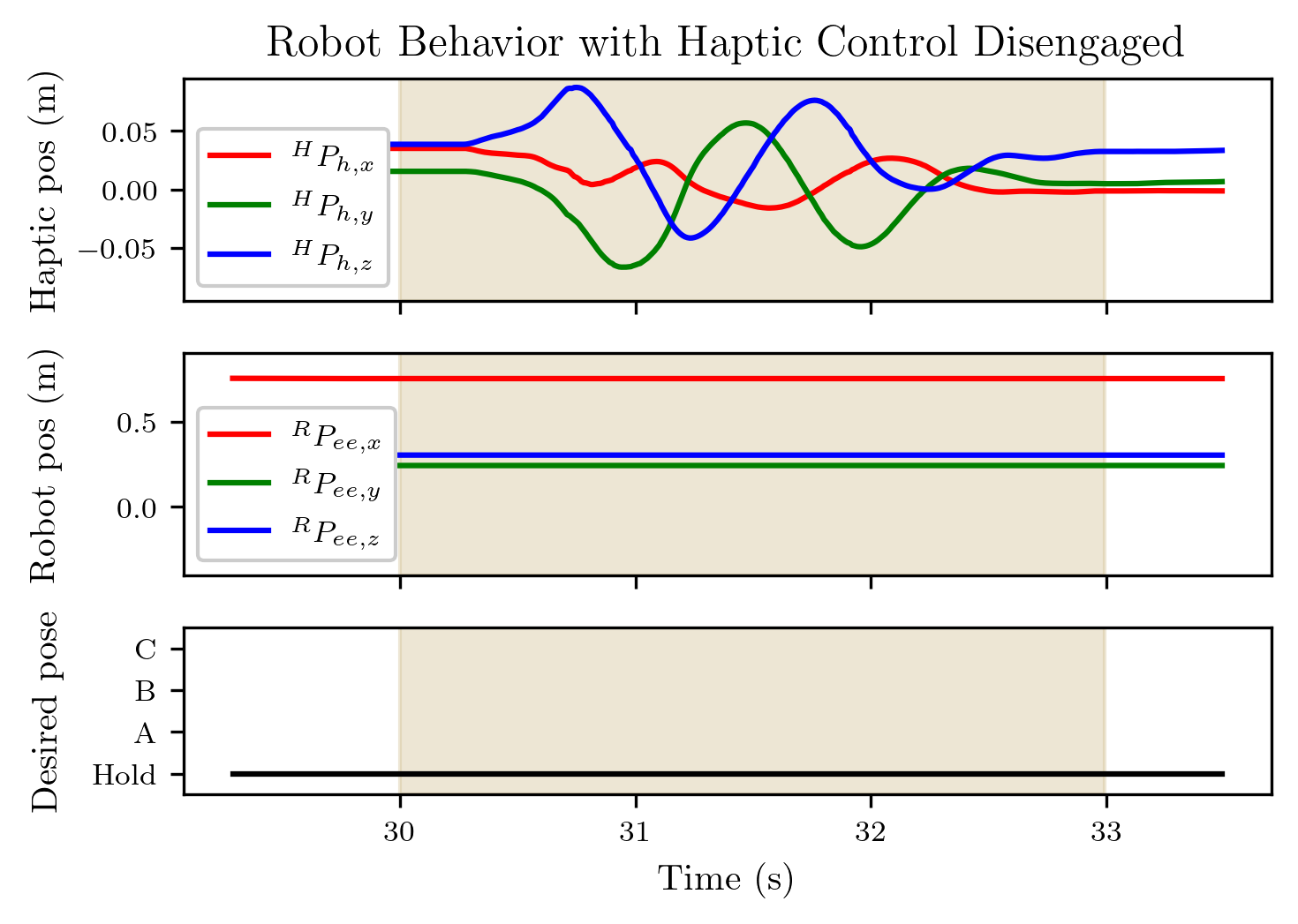}
    \end{subfigure}
    \caption{\textbf{Left:} Demonstration of robot behavior in autonomous (non-shaded and shaded yellow) versus haptic (shaded green) control modes. \textbf{Right:} Zoomed-in view of the robot end-effector disregarding the haptic device position when the haptic control mode is disengaged.}
    \label{fig:switch}
\end{figure*}


\subsection{Real-time Communication} 
We use a real-time operating system to ensure proper communication between the robot, haptic device, hand, force sensor, and vision system. These components are connected to a single computer running Ubuntu 16.04 with the fully preemptible kernel version 4.14.12. Each computer uses an Intel i7 eight-core CPU and seven of the cores are isolated for running the robotic subsystem. 

The isolated cores run the following programs with real-time priority (one program for each core): (1) Controller for computing the desired robot and hand torques, (2) Driver for reading sensor values from the robot and writing desired torques to the robot, (3) Driver for reading sensor values from the hand and writing desired torques to the hand, (4) Controller for computing the desired haptic device forces (for haptic rendering), (5) Driver for reading sensor values from the haptic device and writing desired forces to the haptic device, (6) Driver for reading sensor values from the force sensor, (7) Communication server used to store the sensor and target (if any) values for all subsystem components.
    
The remaining non-isolated core is used to operate the GUI and read sensor values from the vision system. These programs run with non-real-time priority because they do not require real-time communication frequency (as opposed to the remaining subsystem components), and they would otherwise conflict with each other.
    

\section{Results}





The experimental results demonstrate the capabilities of our framework for robot manipulation and the characteristics of the robot primitives used to complete the screwing tasks.

\subsection{Switching Haptic and Autonomous Control} In Fig. \ref{fig:switch}, we demonstrate the robot behavior when switching between autonomous and haptic control modes. The haptic device and robot positions are shown along with the robot's desired autonomous pose. The desired pose can be the robot's current pose (Hold) or one of three preset poses. The haptic device position is expressed in the haptic device base frame, ${}^{H}\!\bm{P}_{h}$, and the robot's end-effector position is expressed in the robot base frame, ${}^{R}\!\bm{P}_{ee}$. When the haptic control mode is engaged, the haptic device position is sent as the desired robot end-effector position with the following affine transformation:

\begin{equation}
    {}^{R}\!\bm{P}_{ee,d} = {}^{R}_{H}\!R  {}^{H}\!\bm{P}_{h} + {}^{R}\!\bm{P}_{h_{rel}} \label{eq:transform}
\end{equation}

\noindent where ${}^{R}\!\bm{P}_{ee,d}$ is the desired robot end-effector position, ${}^{R}_{H}\!R$ is the rotation from the haptic base frame to the robot base frame, and ${}^{R}\!\bm{P}_{h_{rel}}$ is the relative haptic device position from when haptic control was disengaged to when haptic control is reengaged, expressed in the robot base frame. This relative position is initialized as zero.

In the haptic control mode (shaded green), the robot end-effector tracks the haptic device's position. The orientation tracking follows a similar trend.

In the autonomous control mode (non-shaded and shaded yellow regions), the robot's end-effector either holds its current pose or tracks one of three preset poses. In the shaded yellow region, the haptic device position moves while the robot end-effector holds position.

\subsection{Visual Perception and Pose Tracking} 
\begin{figure}[t!]
    \captionsetup{aboveskip=0pt}
    \centering
    \begin{subfigure}{\columnwidth}
      \hfill
      \includegraphics[width =0.96\columnwidth]{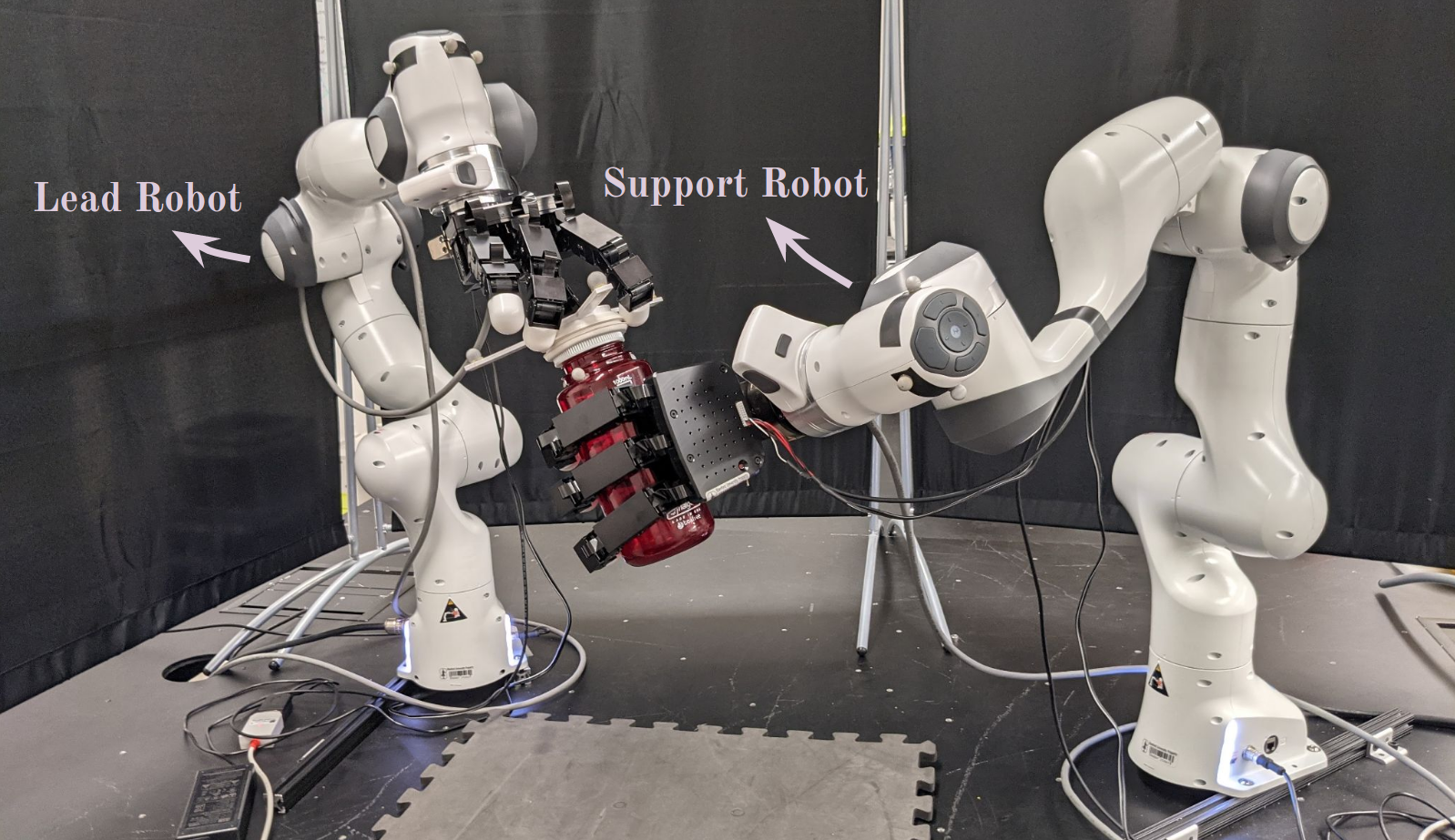}
    \end{subfigure}

    \begin{subfigure}{\columnwidth}
      \centering
        \includegraphics[width =\columnwidth]{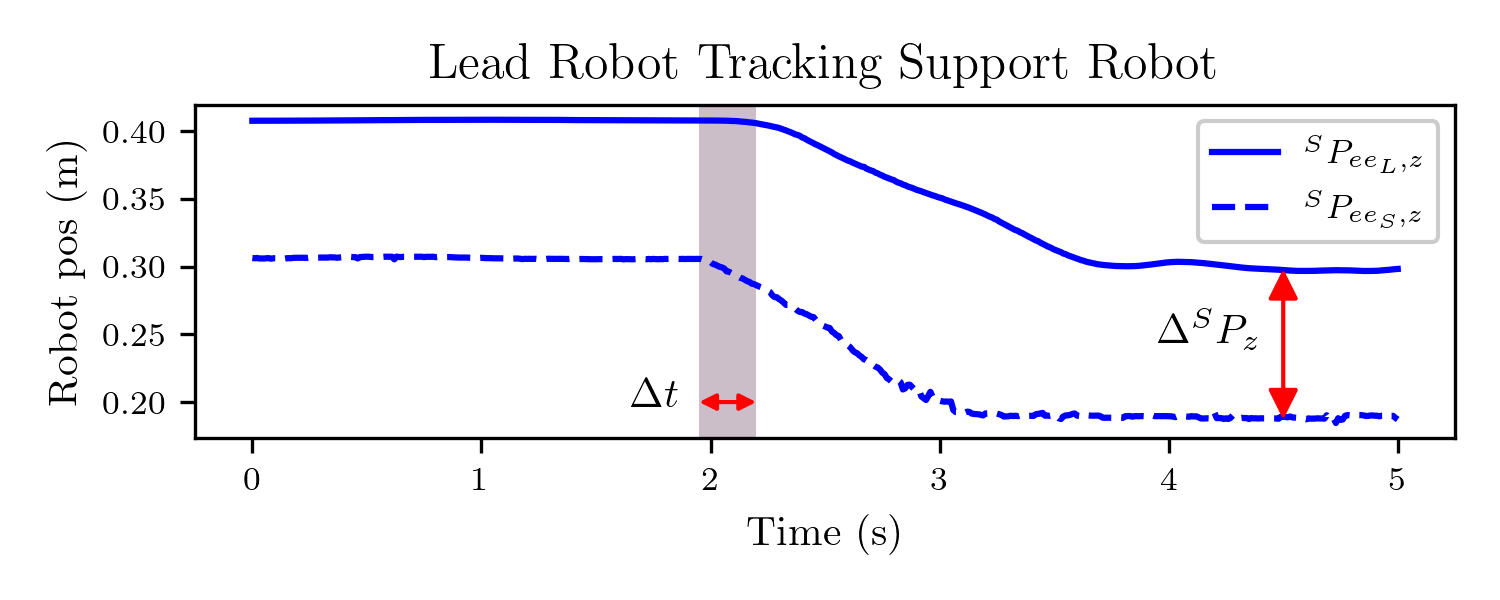}
    \end{subfigure}
    \caption{\textbf{Top:} The support robot holds the bottle. The lead robot holds the cap and is controlled to track the bottle's sensed pose with an offset to avoid making contact. \textbf{Bottom:} The lead robot tracks the support robot position with offsets in time and space.}
    \label{fig:track}
\end{figure}

In Fig. \ref{fig:track}, we demonstrate the go-to-point robot primitive, which is the first step of successful completion of the two-arm bottle screwing task. Here, the lead robot holds the cap and follows the support robot, which holds the bottle. The support robot moves in free space; it can be controlled autonomously or with the haptic device. The bottle is affixed with markers and tracked by the mocap system. The estimated bottle pose is sent as the lead robot's end-effector pose, with a translational offset of $10\,\mathrm{cm}$ along the bottle's longitudinal axis.

The z-position of the lead and support robot end-effectors expressed in the support robot base frame, ${}^{S}\!P_{ee_{L},z}$ and ${}^{S}\!P_{ee_{S},z}$ respectively, are shown in the plot. In this experiment, the bottle's longitudinal axis was aligned with the support robot's base frame z-axis; therefore, the $10\,\mathrm{cm}$ offset shows in the plot. There is also a time offset showing that the lead robot's translational motion lags behind the support robot's translational motion.

\subsection{Full Task Execution} 
\begin{figure}[t]
    \centering
    \includegraphics[width =0.96\columnwidth]{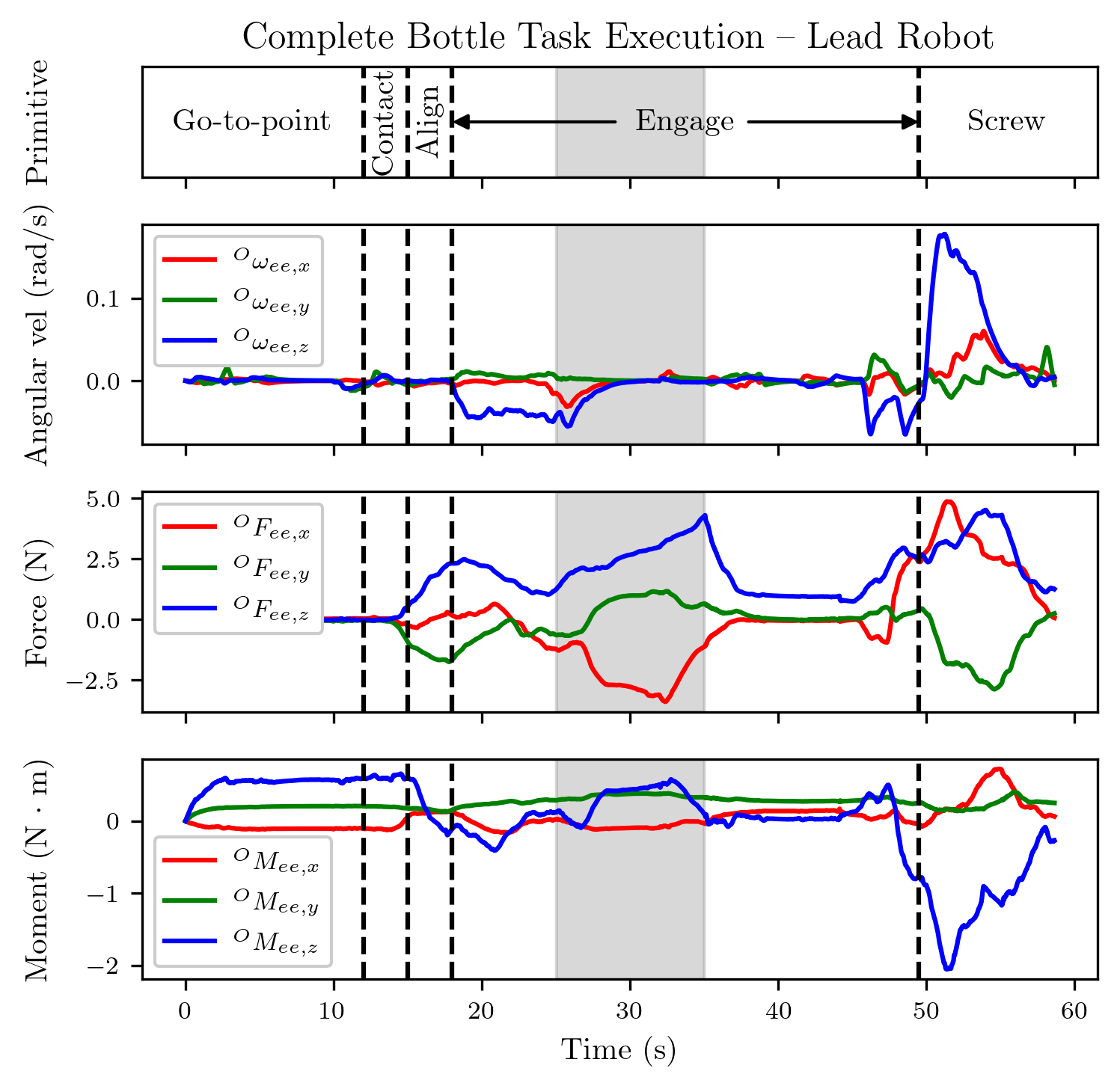}
    \caption{Lead robot state information from a complete execution of the two-arm bottle screwing task. The task is split into five distinct robot primitives. The shaded grey region corresponds to a regrasping action.}
    \label{fig:complete}
\end{figure}

In Fig. \ref{fig:complete}, we show partial state information for the lead robot during a complete execution of the two-arm bottle screwing task. In this task, the lead robot holds the cap while the support robot holds the bottle. The goal of the task is to screw the cap onto the bottle.

The relevant state information is the angular velocity of the lead robot's end-effector expressed in the object (bottle) frame, ${}^{O}\!\bm{\omega}_{ee}$; the lead robot's sensed force expressed in the object frame, ${}^{O}\!\bm{F}_{ee}$; and the lead robot's sensed moment expressed in the object frame, ${}^{O}\!\bm{M}_{ee}$. The task is split into five unique robot primitives: go-to-point, (make) contact, align, engage (the threads), and screw.

In the go-to-point primitive, the lead robot moves in free space to match the bottle's pose with a fixed translational offset along the bottle's longitudinal axis, ${}^{O}\!z$. This behavior is shown in the previous results section. Once the target pose is reached, the lead robot transitions to the next primitive.

In the make contact primitive, the lead robot is controlled to match the bottle's (1) orientation and (2) position in the plane normal to ${}^{O}\!z$. The lead robot is further controlled to maintain a desired force along ${}^{O}\!z$.

Once the desired force is sensed and maintained along ${}^{O}\!z$, the lead robot transitions to align the cap with the bottle. Here, the lead robot is controlled to (1) match the bottle's position in the plane normal to ${}^{O}\!z$, (2) maintain the desired force along ${}^{O}\!z$, and (3) match the bottle's orientation about ${}^{O}\!z$. The lead robot is also controlled to achieve zero-moment about the plane normal to ${}^{O}\!z$. The alignment is complete when the lead robot senses and maintains zero moment while maintaining the desired force along ${}^{O}\!z$.

Next, the lead robot transitions to engage the bottle's threads with the cap. The lead robot is controlled to maintain its position and orientation, except for its orientation about ${}^{O}\!z$. The lead robot rotates incrementally about $-{}^{O}\!z$ (i.e. in the loosening direction) until a joint limit is reached.

Once a joint limit is reached in the engage primitive, the lead robot starts the regrasping action, which is highlighted in the shaded grey region of Fig. \ref{fig:complete}. In order to regrasp, the lead robot lets go of the cap and rotates incrementally about ${}^{O}\!z$ before grasping the cap again. While the cap is not grasped, the sensed force is not rendered to the haptic device since the force sensor is calibrated with the cap grasped.

After regrasping the cap and engaging the bottle threads, the lead robot transitions to the screw primitive. To screw the cap onto the bottle, the lead robot rotates the cap about ${}^{O}\!z$ until a sufficiently large moment is sensed about $-{}^{O}\!z$.

Elly successfully completed over 80\% of the 40 experiments focused on task completion listed in Table \ref{tab:experiments}. Most of the task failures were due to dropping the cap during the task without being able to regrasp it, which was not a focus of this paper.

%% file: 5-Conclusions.tex
\section{Conclusions and Future Work}
\label{sec:conclusions}

In this work, we presented Elly: a system that combines robot autonomy in the form of model-based primitives and human inputs through a haptic device to ensure task completion during contact manipulation tasks. Furthermore, the haptic device and graphical interface allow the operator to collect data that can be used to improve the robot's autonomous behaviors over time. Our experimental results validate Elly on a complex hardware setup involving two 7-DOF manipulators equipped with 4-finger grippers, two haptic devices, two force sensors, and a mocap system. 

We have identified several improvements to be addressed in future work. First, in terms of data collection, it would be valuable to automatically synchronize the GUI's logging start time with the start of the autonomous controller. This would make the data easier to interpret during post-processing. Second, on the controller side, it could be useful to provide more flexibility to the user regarding the degrees of freedom they want to control haptically. For example, it is sometimes helpful to have the robot autonomously hold position while the operator only controls orientation. Finally, we would like to take the light bulb task one step further and attempt to perform the installation inside a lamp where the socket is occluded using only contact data. 

In summary, Elly effectively enables manipulators to learn safe, modular behaviors that work on a complex hardware setup and take human input to incorporate knowledge about new scenarios over time. We believe Elly could be used by other researchers to collect their own datasets as well as inform their design decisions when building a bimanual system for learning autonomous behaviors.


